# Interval type-2 Beta Fuzzy Near set based approach to content based image retrieval

Yosr Ghozzi, Nesrine Baklouti, Hani Hagras, Mounir Ben Ayed, and Adel M. Alimi, Member, IEEE

**Abstract**— In an automated search system, similarity is a key concept in solving a human task. Indeed, human process is usually a natural categorization that underlies many natural abilities such as image recovery, language comprehension, decision making, or pattern recognition. In the image search axis, there are several ways to measure the similarity between images in an image database, to a query image. Image search by content is based on the similarity of the visual characteristics of the images. The distance function used to evaluate the similarity between images depends on the criteria of the search but also on the representation of the characteristics of the image; this is the main idea of the near and fuzzy sets approaches. In this article, we introduce a new category of beta type-2 fuzzy sets for the description of image characteristics as well as the near sets approach for image recovery. Finally, we illustrate our work with examples of image recovery problems used in the real world.

## I. Introduction

The number of daily-generated images by websites and personal archives are constantly growing. chives reaching unimaginable sizes. [1], [2]. Indeed, the effective management of the rapid expansion of visual information has become a major problem and a necessity for strengthening visual search technique based on visual content [3]. This necessity is behind the emergence of new visual search techniques based on visual content. It has been widely identified that the most efficient and intuitive way to research visual information is based on the properties that are extracted from the images themselves. Researchers from different communities ("Computer Vision" [4], "Database Management", "Man-machine Interface", "Information Retrieval") were attracted by this field. Since then, the search for images by content has developed quite rapidly. The intuitive idea of "any system that analyzes or automatically organizes a set of data or knowledge must use, in one form or another, a similarity operator whose purpose is to establish similarities or the relationships that exist between the manipulated information". To find the images that most resemble an example image or to group them together [5], we must be able to measure the similarity (or dissimilarity) of the images [6]. The similarity measures really useful for image search are those that are close to human perception. However, many works attempt to draw inspiration from the human visual system to propose more effective measures. Search content (or CBIR for content-based image retrieval) [7] consists of extracting images of the large dimensional vectors called descriptors and associating to them a similarity measure.
The aim of this method is to reduce the notion of visual similarity between images to a simple notion of proximity between the descriptors. Finding images similar to a query image is equivalent to searching for neighbors closest to the descriptor of the query image in the description space. The nearest neighbors allow to designate the images most similar images to the query image. This new approach has the major consequence of querying a database of images directly from its visual content. For several years many works on the search for images based on the visual content have been born. Among them, Peters in presented an approach using near sets and tolerance classes. This method is developed in the context of perceptual systems [8], where each image or part of an image is considered as perceptual objects [9]. "A perceptual object is something presented to the senses or knowable by the human mind" [10]. Actually, among the human body's mechanisms, visual perception represents one of the most complex ones.

Our brain has the capacity to analyze intricate scenes in a split second. Today, powerful image processing software is available to the public the manipulation or modification of images. The image processing methods merely associate with each image a vector (or vectors) of characteristics calculated on the basis of the so-called 'low level' image characteristics (color, texture, shape, etc.). The querying of an image database is then carried out by introducing a query image into the system and comparing the characteristics thus calculated using a similarity measure. In information systems, some features or attributes may not offer distinctive characteristics for a object (set). Therefore, the assertion that some features (attributes) of one object partially or completely match those of another does not imply that they are tightly related. In other words, the only assertion that fits is that one object on the distinguishing features partially or completely matches those of another object, or two objects are closely related. In this paper, we extend near sets based on AFS fuzzy description logics, in which the closeness (nearness) of objects, if and only if they have similar fuzzy descriptions. The aim of this work is to take a step towards the real case (the search of images by the content for the detection of similar images visually), on the one hand, and to evaluate the matching algorithm and the similarity measure we used in the CBIR context. There are various areas to work in for the improvement of the content based image retrieval system. It has already been discussed that the existing techniques may be used to improve the quality of image retrieval and the understanding of user intentions.

An approach that combines two different approaches to image retrieval, together with the active use of a near set approach and the fuzzy set has been proposed. The use of the hybrid approach of processing image as feature vector fuzzy of the regions to match images can give better results.

The present work has two aims, the first of which is to take a step towards the real case (image of the real world) and the second is to evaluate the matching algorithm and the similarity measure that we used. It is structured around six main sections: After the introduction, section two presents related research works to the field of image-based image retrieval that are realized by different techniques and methods of type 2 fuzzy sets and near sets. Section three displays the theoretical foundations of the different techniques of fuzzy sets and near sets. AS for section four, it is devoted to the presentation of the sought objectives pursued and to the adopted research methodology. The experiments and results are presented in section five. Finally, a conclusion is presented.

## II. RELATED WORK

In this section, a representative review of some systems using Interval Type-2 Fuzzy set and Near set in Content-Based Image Retrieval (CBIR) is presented.

In [11], the authors have given a practical implementation of the flow graphs induced by a perceptual system, defined with regard to digital images, to perform CBIR. The results are generated using the SIMPLicity dataset, and compared with the near-set based tolerance nearness measure (tNM).

Furthermore [12] has shown that tolerance near sets can be used in a concept-based approach to discover correspondences between images, from an application of anisotropic (direction dependent).

Besides, Rahman in [7], has used the fuzzy sets with a similarity measure. In this work, the image similarity measure is improved through a fuzzyfication of regions importance and inter-region similarity. The region-based image comparison is defined as two images that are usually compared in terms of the sum of the Euclidean distances among their regions. The utilization of fuzzy concepts of the size and shape features of the regions; these two functions impose additional constrains on similarity measure that helps to improve the image retrieval results.

Gupta proposed in [14], a new fuzzy-based approach with Genetic Algorithm-based to develop a hybrid similarity measure that overcomes the limitations of extensively used similarity measures, such as Cosine, Jaccard, Euclidean and Okapi-BM25 along. This approach uses fuzzy rules to infer the weights of different similarity measures.

In [15], a new semantic approach for CBIR supported by a parallel aggregation of content-based features extraction (shape, texture, color) using fuzzy support decision mechanisms has been presented. Fast Beta Wavelet Network modeling and Hue moments are the rudiments of shape features. The texture descriptor is based on Energy computing at various decomposition levels.

In [16], the author use T1 and T2 fuzzy models in a supervised image segmentation algorithm was proposed, to ameliorate the performance of the final model. Qualitative and quantitative analyses have demonstrated that it has better accuracy than other common techniques when using both synthetic image datasets and panchromatic images.

Nonetheless, Castillo in [18] has shown that type-2 fuzzy sets outperform both traditional image processing techniques as well as techniques using type-1 fuzzy sets, and provide the ability to handle uncertainty when the image is corrupted by noise.

In [19], the authors have proposed a approach with encouraged performance. He extracted Fuzzy-Object-Shape information in an image for provides a measure of closeness of this object of interest with well-known shapes.

In [20], El Adel have developed a texture image retrieval system that learns the visual similarity in terms of class membership using multiple classifiers. The way proposed approach combines the decisions of multiple classifiers to obtain final class memberships of query for each of the output classes are also a novel concept.

In [21], a novel approach has been proposed to retrieve digital images using texture analysis techniques to extract discriminant features together with color and shape features.

However, in [22], the most similar highest priority (MSHP) principle is used for matching of textural features and Canberra distance is utilized for shape features matching. The retrieved image is the image which has less MSHP and Canberra distance from the query image.

## III. PRELIMINARIES

### A. Near sets basis

Near sets gather disconnected sets to each other [23]. Disjoint sets are assembled whenever similarities between the objects in the sets are observed. As similarity is determined by comparing lists of object feature values, each of which denotes an object's description. A feature is essentially a characteristic of the aspect of what the perceptual items are made up of; the perceived items. A probe function is an actual evaluated function that represents the characteristics of the perceived items [24]. Within the framework of the Near Set Theory, the items of the perceived field are usually presented on the basis of the chosen probe functions. This implies that the role of the probe function is to assess the characteristics of the perceived of perceived items related to with a group of probe functions. Indeed, a perceptual item in a conceptual system can be described as follows. Let $O$ be represent a set of perceptual objects, and B denote a set of real-valued functions, denoted probe functions, representing object features, and let $\varphi(x) \in B$, where $\varphi_i(x): O \to R$. Similarly, the functions representing object features offer a vector comprising measurements (returned values) for an object description, linked to each functional value $\varphi_i(x)$ for $x \in X$, where $|\varphi| = l$; i.e. the description length is $l$.

**Object Description**:
In what fellows, the relationship between objects is identified by the probe functions in B. Our senses are defined to probe functions. The tolerance space definition, a specific tolerance relation [24] is given by:

**Definition:**
Let $(O,F)$ be a perceptual system and let $\varepsilon \in R_0^+$ (real). For every $B \subset F$ the weak tolerance relation $\cong_{B,\varepsilon}$ is defined as:
$$\cong_{B,\varepsilon} = \{(x,y) \in O \times O | \exists \varphi_i \in B \cdot \|\varphi_i(x) - \varphi_i(y)\|_2 \leq \varepsilon\} \quad (1)$$

It is worthy to note that although the relation $\cong_{B,\varepsilon}$ is symmetric and reflexive, it is not transitive which is very important in finding near sets, as it typifies characterizes tolerance classes within a threshold $\varepsilon$.

Lastly, the concept of near sets is established on the propositions requiring neighborhoods and tolerance classes.
These concepts are described by: Let $(O,F)$ be a perceptual system, and let $x \in O$. For a set $B \subset F$ and $\varepsilon \in R_0^+$, a neighborhood is:
$$N(x) = \{y \in O : x \cong_{B,\varepsilon} y\} \quad (2)$$

The separated classes that incorporate similar items are said to be neighbors. Actually, similitude is arithmetically identified through an item description. The Near Set Theory represents a proper ground for the determination, the comparison, and the measurement of the similitude of the items through the description of their characteristics. The near sets come out when the feature vectors are identified to describe and distribute the similarities between the sample components. Therefore, the elements that have similarities between their characteristics are supposed to be perceptually near one another. The classes of these items that are obtained from the set separation give more information and show some forms of interests.

Neighboring sets are identified by a tolerance relationship based on description. Similarity measures represent one of the necessary components of image databases. They allow checking whether two images are duplicate, alike in some measure or totally different. Many methods of expressing similarity exist, depending on the method by which the equivalence between the images is assessed. Therefore, this research work seeks to detect the usage manner of these measurements to perform automatic search, particularly in the context of the search by visual regions. The basis for the application of the Near Set Theory is the notion of nearness between two sets [8]. The tolerance nearness measure is a quantitative approach that determines the extent to which the near sets take after one another. This approach was created due to a need for solving the problem of the Near Set Theory to the practical applications of image equivalence [7]. The idea behind the nearness measure of Henry and Peters is sought after as the level of similarity between two variables by eliminating the existing correlation between the sets of variables, called the tolerance classes.

The correspondence measures can be clustered into equivalent classes of measures. The tolerance nearness measure between two sets X and Y is based on the concept that equivalent classes formed from objects in the union $Z = X \cup Y$ should be uniformly divided between X and Y if these sets are similar.

**Definition:**

The tolerance nearness measure : Let $(O,F)$ be a perceptual system and let $\varepsilon \in R_0^+$ (real $\Re$). For every $B \subset F$. Moreover, let X and Y be two disjoint sets. A tolerance nearness measure between two sets X and Y is determined by:

$$tNM_{\cong_{B,\varepsilon}}(X,Y) = 1 - \left(\sum_{C \in H_{\cong_{B,\varepsilon}}(Z)} |C|\right)^{-1} \cdot \sum_{C \in H_{\cong_{B,\varepsilon}}(Z)} |C| \frac{\min(|C \cap X|, |C \cap Y|)}{\max(|C \cap X|, |C \cap Y|)}$$

where $H_{\cong_{B,\varepsilon}}(Z)$ is tolerance classes.

*B. Type 2 Beta Fuzzy basis*

In the previous section, it was shown how tolerance relation can be used in modeling the existing imprecision in human visual perception of the physical world. Tolerance relations can be used as a basic framework for modeling this tolerance level of difference in description (physical feature). The existing tolerance in overlooking small changes in visual appearances is one aspect of the human perception. However, it is not clear if there is a sharp crisp threshold for this tolerance. In fact, the exact equality of descriptions is not necessary to consider two objects similar. There is always an admissible level of error in comparing objects by their description. Incorporating the concept of tolerance is not only permissible but also necessary to arrive at approximate solutions of problems in real world. Therefore, the transition from "similar" to "dissimilar" in human mind is gradual not abrupt. There is no boundary between similar and dissimilar and it is just the matter of degree of similarity, and thus the intrinsic fuzziness in this concept. A fuzzy relation is a solution for incorporating the concept of fuzziness or imprecision in the definition of similarity.

The objective of this section is to introduce a more general approach based on fuzzy tolerance relations that can address all the above aspects in defining the similarity between objects or sets of objects. The information provided by the image items and the similitude between them. The comparison of object descriptions is the building stone of defining the measure that disjoint sets look like each other. Based on their similar descriptions, the set of object are clustered. These groups of similar objects can reveal similar patterns and information about the objects of interest in the disjoint sets. Near set theory concentrates on the sets of perceptual objects with comparable descriptions. Tolerance near sets are determined by a tolerance relationship based on description. Tolerance relationship gives an intransitive idea about the world. In fact, tolerance near sets give an appropriate foundation for the majority of the solutions that are moderately valid. These sets are required when handling with the world's problems and applications [25]. In other words, tolerance near sets are used as a fundament for a qualitative method to evaluate the resemblance between the items without the need for these items description. A definition of the content of the sets evinces that any item in the Near Set Theory includes perceptual items. This alludes to anything in the physical world that has characteristics that are likely to be perceived by the senses since they can be evaluated and assumed by the mind. Indeed, a feature is a characteristic of the aspect of what the perceptual items are made up of; the perceived items. A perceptional system is a group descriptions are uncertain and imprecise. We manage these ambiguities by using a fuzzy approach. The most important part of the fuzzy logic theory is the modification of the membership values by means of various fuzzy techniques, once the image descriptors has been transformed from the crisp value plane to that the plane of the membership values by this stage of fuzzification. The fuzzification plays a major role in handling the data in fuzzy environment.

A fuzzy set is a collection of objects in connection with the expression of uncertainty of the property characterizing the objects by grades from interval between 0 and 1 [26].

**Definition:**

A fuzzy subset of a set S is a realization μ of S in [0, 1]. For all p in S, μ(p) is called degree of membership of p in [0, 1]. The fuzzification is based on certain membership function. The Triangular, Trapezoidal or Gaussian shapes are the most commonly used forms for membership functions in fuzzy set. The beta distribution is seen as a suitable model in data analysis because it provides a wide variety of distributional shapes over a finite interval [27]–[30]. The exploitation of such function proves higher performance as compared with the other types of functions due to its universal approximation. Unfortunately, the beta distribution is not easily understood and its parameters are not easily estimated. This is a family of laws of continuous probabilities, defined in [0,1], parametrized by two shape parameters, typically denoted α and β. Admitting a great variety of forms, it allows to model many finite support distributions.

**Definition:**

A Beta function in one-dimensional case, is given by the following equation:
$$Beta(x; \alpha, \beta, x_{min}, x_{max}) = \begin{cases} (\frac{x - x_{min}}{x_{center} - x_{min}})^\alpha \cdot (\frac{x_{max} - x}{x_{max} - x_{center}})^\beta, & if\ x \in ]x_{min}, x_{max}[ \\ 0, & else \end{cases} \quad (4)$$

In which $\alpha, \beta, x_{min}$ and $x_{max} \in R$ such that $\alpha, \beta > 0$ and $x_{max} > x_{min}$. $x_{center} = (\alpha \cdot x_{max} + \beta \cdot x_{min})/(\alpha + \beta)$ is the center of Beta function and $\sigma = x_{max} - x_{min}$ its width.

Indeed, the standard triangular distribution is a special case of beta distribution, with modification of only two parameters; left parameter $x_{min} = a$ (the lower mode) and right parameter $x_{max} = b$ (the upper mode), with $\alpha = \beta = 1$.

$$triangular(x; a, b) = \begin{cases} \frac{x-a}{b-a}, & a \leq x < b \\ 0, & else \end{cases} \quad (5)$$

The triangular distribution is also a special case of the trapezoidal; It is missing the constant stage.

The trapezoidal distribution is a special case of the beta distribution; left parameter $x_{min} = a$ (the lower mode) and right parameter $x_{max} = d$ (the upper mode), with $\alpha = \beta = 1$.

$$trapezoidal(x; a, b, c, d) = \begin{cases} \frac{x-a}{b-a}, & a \leq x < b \\ 1, & b < x < c \\ \frac{d-x}{d-c}, & c < x \leq d \\ 0, & else \end{cases} \quad (6)$$

Thus, for the Gaussian function $Gaussian(x; \mu, \sigma)$, there exists a Beta function that approximates for any given precision $\varepsilon$, for more details see [28].

$Beta(x; \alpha, \beta, x_{min}, x_{max}) - Gaussian(x; \mu, \sigma) < \varepsilon$ for any $\varepsilon \in R$. So Eq. 4 leads to:

$$Beta(x; x_{center}, \sigma, \alpha, \beta) = \begin{cases} \left(1 + \frac{(\alpha+\beta)(x-x_{center})}{\sigma\alpha}\right)^\alpha \cdot \left(1 - \frac{(\alpha+\beta)(x_{center}-x)}{\sigma\beta}\right)^\beta, \\ \quad if\ x \in \left]x_{center} - \frac{\sigma\alpha}{\alpha+\beta}, x_{center} + \frac{\sigma\beta}{\alpha+\beta}\right[ \\ 0, else \end{cases} \quad (7)$$

Consequently, we can see the beta approach more easily by modifying these parameters, transformed by the many distribution of function.

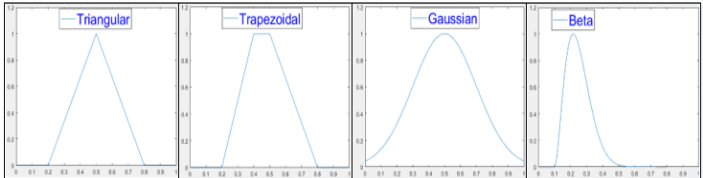

Fig. 1. Different forms of membership functions Fuzzy set

Figure 1 illustrates some distribution of its parameters, different forms of membership functions: triangular, trapezoidal and gaussian distribution, which are alternative solutions are included for beta distribution. The main importance of the Beta function lies essentially on its capacity to approximate many usual functions. The transition from an ordinary set to a fuzzy set is the direct consequence from the indeterminacy of the value of the membership of an element to a set by 0 or 1. Similarly, when we cannot determine the membership functions (MF) fuzzy by real numbers in [0; 1], we use then the type-2 fuzzy sets. For this, we can consider type-1 fuzzy sets to be a first-order approximation of uncertainty and type-2 fuzzy sets a second order approximation.

An interval type-2 fuzzy set (IT2FS) [31] is characterized by a fuzzy membership function, i.e. the membership value (or membership grade) for each element of this set is a fuzzy set in [0; 1], unlike a type-1 fuzzy set where the membership grade is a crisp number in [0; 1].

**Definition:**

An interval type-2 fuzzy set, denoted $\tilde{A}$, is characterized by a type-2 membership function $\mu_{\tilde{A}}(x,u)$, where $x \in X$ and $J_x \subset [0,1]$, i.e.,

$$\tilde{A}(x) = \int_{x \in X} \int_{u \in J_x} \mu_{\tilde{A}}(x, u)/(x, u)\ J_x \subset [0,1] \quad (8)$$

Where $0 \leq \mu_{\tilde{A}}(x, u) \leq 1$ and $J_x$ is the closure of $\mu_{\tilde{A}}(x, u) > 0$. For any given $x \in X$.

$$\mu_{\tilde{A}}(x) = \int_{u \in J_x} \mu_{\tilde{A}}(x, u)/u \quad (9)$$

is called a second membership function, clearly, it is a type-1 fuzzy set. An IT2FS is represented by a bounded region limited by two MFs, corresponding to each primary MF (which is in [0; 1]). The Uncertainty in the primary MF consists of the union of all MFs. This Uncertainty represents a bounded region that we call the Footprint of Uncertainty (FOU). This region represents a complete description of an IT2FS. IT is delimited by two MFs noted the Upper MF (UMF), which is denoted $\bar{\mu}_{\tilde{A}}(x)$ and the Lower MF (LMF), which is denoted $\underline{\mu}_{\tilde{A}}(x)$, i.e.,

$$FOU(\tilde{A}) = \bigcup_{x \in X} u \in J_x;\ J_x = \left[\bar{\mu}_{\tilde{A}}(x), \underline{\mu}_{\tilde{A}}(x)\right], \forall x \in X \quad (10)$$

$$\bar{\mu}_{\tilde{A}}(x) \equiv \overline{FOU(\tilde{A})}\ \underline{\mu}_{\tilde{A}}(x) \equiv \underline{FOU(\tilde{A})}\ \forall x \in X \quad (11)$$

In this work, the Beta basis function is chosen for the modeling of the IT2FS. Hence, a Beta primary MF having an interval valued secondary MF is adopted and termed the Interval type-2 Beta MF. The proposed beta MF has a uncertain center $x_{center}$, an fixed width $\sigma$ and fixed form parameters $\alpha$ and $\beta$. However, the upper and lower membership functions can be expressed by respectively:

$$\bar{\mu}_{\tilde{A}}(x) = Beta(x; x_{center1}, \sigma, \alpha, \beta) \quad (12)$$
$$\underline{\mu}_{\tilde{A}}(x) = Beta(x; x_{center2}, \sigma, \alpha, \beta) \quad (13)$$

Where $x_{center1} = x_{center} \cdot \alpha$ and $x_{center2} = x_{center} \cdot \beta$.

Figure 2 illustrates some examples of type-2 Triangular MF, type-2 Trapezoidal MF, type-2 Gaussian MF and interval type-2 Beta MF.

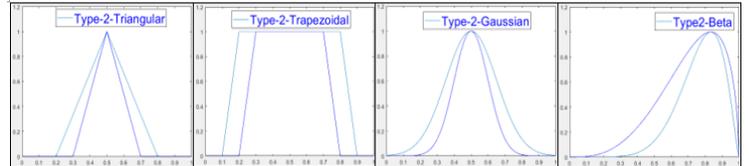

Fig. 2. Different forms of membership functions Type-2 Fuzzy set

*C. Fuzzy Near Sets*

As previously mentioned by Peters in [32], [33], a fuzzy set X is a near set relative to a set Y if the grade of membership of the objects in sets X, Y is allocated to each object by the same membership function $\varphi$ and there is a least one pair of objects $x, y \in X \times Y$ such that $\|\varphi(x) - \varphi(y)\|_2 \leq \varepsilon$, i.e., the description of x is similar to the description y within some $\varepsilon$.

**Proposition:**

Peters proposed in [33]; Two Fuzzy sets $(X, \varphi), (Y, \varphi)$, are weakly near sets if, and only if there exists at least one tolerance class $x/\cong_{\varphi,\varepsilon}$ in $(X, \varphi)$, and $y/\cong_{\varphi,\varepsilon}$ in $(Y, \varphi)$, such that $x/\cong_{\varphi,\varepsilon} \blacktriangleright \blacktriangleleft_{\varphi,\varepsilon} y/\cong_{\varphi,\varepsilon}$

This is a fuzzy near set model that has been used in the proposed algorithm in [25]. Let be two images as a source image S and a target image T of the same object. The image is divided into blocks (sub-image) in a uniform way, in which each sub-image is roughly treated as an object. In the near sets sense, each sub-image is a perceptual object and each object description comprises the values obtained from image processing techniques on the sub-image. The membership of each feature is calculated using fuzzification function. Based on the fuzzy feature representation of images, illustrating the similarity between images has become an issue of finding those between fuzzy features.

## IV. A INTERVAL-TYPE-2 BETA FUZZY NEAR METHOD (T2FNM) IN IMAGE RETRIEVAL SYSTEM

This section describes the image retrieval system which introduce the near fuzzy set in resemblance between images. Based on fuzzy feature representation of images, characterizing the similarity between images has become an issue of finding those between fuzzy features. We follow the steps described later in our system according to the diagram Figure 3.

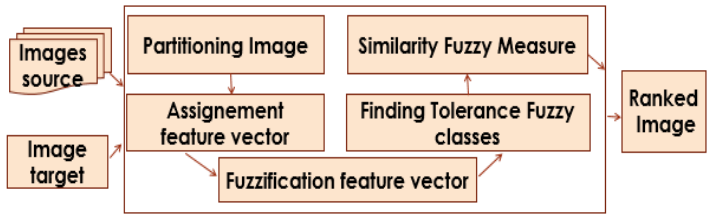

Fig. 3. Architecture of IT2FNM based System

### A. Pre-processing image database

*1)* Partitioning Image:

This research work is based on a set of theoretic approaches to image analysis in which each image is viewed as a set of visual elements (or more generally, describable objects). Each visual element can be just a pixel, a pixel and its surrounding pixels or any part of the image that can be visually perceived and described. It is due to a practical and physical reason that this research has been undertaken using a visual element rather than a single pixel. From a practical standpoint, it is easier to consider a small patch of adjacent pixels as a unit of visual perception and therefore decreasing the amount of information required to represent the image as it is perceived by a human. From a physical viewpoint, it is known that images are not seen in a pixel-based resolution and the local perception of the image is formed by a group of pixels. That is why we decided to decompose the images of our base into blocks of fixed size. Indeed, the image is uniformly divided into blocks is nearly assimilated to a sub-image.

The size of these blocks is intelligently chosen to be as small as possible to represent local details in an image, on the one hand, and as large as possible to limit the number of visual elements for the sake of speed in the algorithm, on the other. To do so, a study has been undertaken in [34] to determine the most interesting window size for an efficient extraction of image primitives. An image of size 256 by 384 is decomposed into fixed-size blocks as a square sub-image of size 13 by 19 pixels and each block represents an object of the image.

*2)* Description Image:

In Near Set theory, a visual element represents something in the physical world and thus it can be perceived and described. It is possible to describe the element through a set of characteristics (features). A visual element (as described in the previous section) is a sub-image that can be perceived and described by color, shape or texture (probe function). However, this step is to automatically extract significant visual characteristics from the image and store them in a digital vector called visual descriptor. The choice of the extracted characteristics is often guided by the will of invariance or robustness with respect to transformations of the image.

*3)* Fuzzification features:

In this step, the proposed contribution is presented by introducing the notions of the fuzzy logic. The fuzzification consists of characterizing the features of the image by the linguistic variables. It is therefore a transformation of the real inputs into a fuzzy part defined on a representation space linked to the input. This representation space is normally a fuzzy subset. In this representation, each sub-image is related to a fuzzy feature that allocates a value (between 0 and 1) to each feature vector in the feature space. The value named degree of membership exemplifies the degree of membership with a matching feature vector which characterizes the sub-image, and thus modeling its uncertainties. Building or choosing a suitable membership function is an application-dependent problem. Some most commonly used prototype membership functions are triangular, gaussian, trapezoidal, Type-2, and Beta functions (as described in the previous section). Two factors are considered when we choosing the membership function for the proposed system: retrieval accuracy and computational intensity for assessing a membership function. In our case, the most suitable form is the beta type-2 form. This form was chosen empirically from comparative tests with trapezoidal, triangular, Gaussian and beta forms. This choice is compatible with the results obtained for other application cases.

### B. Computing the Tolerance Fuzzy Relation

The next step is the search step consisting in matching the descriptor vector of the query image proposed by the user with the descriptors of the database using a distance measurement to obtain a satisfactory matching in near sets sense.

Define a tolerance fuzzy relation $\cong_{\varphi,\varepsilon}$ between feature vectors based on a tolerance level of error $\varepsilon$ to represent similarity in the sub-image level. Two visual elements x and y are similar to each other if the above distance between feature fuzzy vectors $d(\varphi(x), \varphi(y))$ (d Euclidean distance) is smaller than the tolerable level of error threshold. A classical relation R defined on a set X is a subset of $X \times X$ where any of the $\hat{R}$ elements of the Cartesian product has a crisp degree of membership (0 or 1) in the set R. Similarly, a fuzzy relation $\hat{R}$ defined on a crisp set X is a 'fuzzy set' is defined as follows where the membership function represents degree of membership of each pair of elements in the relation (i.e. the degree to which, the elements are related to each other).

$$\hat{R} = \{((x,y), \mu_{\hat{R}}(x,y)) | (x,y) \in X \times X, \mu_{\hat{R}}(x,y) \in [0,1]\} \quad (12)$$

Furthermore, many of the conventional concepts in set theory can be fuzzified. A conventional equivalence relation is a relation that is reflexive, symmetric and transitive.

**Definition:**

Fuzzy Equivalence Relation [35] Let $\hat{R}$ be a fuzzy relation defined on X using the membership function $\mu_{\hat{R}}(x,y)$. $\hat{R}$ is a fuzzy equivalence relation if and only if it has all the following properties:

- Reflexivity: $\forall x \in X, \ \mu_{\hat{R}}(x,x) = 1$

- Symmetry: $\forall x, y \in X, \mu_{\hat{R}}(x,y) = \mu_{\hat{R}}(y,x)$
- Transitivity: $\forall x, y, z \in X, \mu_{\hat{R}}(x,z) \geq \mu_{\hat{R}}(x,y) * \mu_{\hat{R}}(y,z)$

where * represents a triangular norm (t-norm). A t-norm is a commutative, monotonic and associative binary operation defined on [0, 1] [0, 1] into [0, 1]. A simple common example of such function is the minimum function also named as Godel t-norm.

The main motivation for using fuzzy set theory in the definition of similarity measures in this paper is to allow a more humanistic natural-language compatible form of distance measures between pairs of images. Humans do not use numbers to express similarity between images.

Instead, human-judged similarities are expressed in terms of natural language expressions like identical, very similar, partially similar, not similar, etc. Actually, what is meant by very similar (for example) is highly subjective and also depends on the context.

**Definition:**

Let $O$ a set of describable objects, $B$ a set of probe functions and $\varphi_B$ is the set of feature vectors. Suppose $\|.\|_2$ is a distance function on $(\varphi_B, d)$. Let $\varepsilon < \varepsilon' \in R$. A perceptual fuzzy tolerance relationship $\cong_{\varphi,\varepsilon}: O \times O \to [0,1]$ is defined as follows:

$$\cong_{\varphi,\varepsilon} = \begin{cases} = 1 & if \|\varphi_B(x),\varphi_B(y)\|_2 < \varepsilon \\ \frac{\varepsilon' - \|\varphi_B(x),\varphi_B(y)\|_2}{\varepsilon' - \varepsilon} & if \varepsilon < \|\varphi_B(x),\varphi_B(y)\|_2 < \varepsilon' \\ = 0 & otherwise \end{cases}$$
(13)

Figure 4 shows the transition gradual in a fuzzy tolerance relation of the similarity.

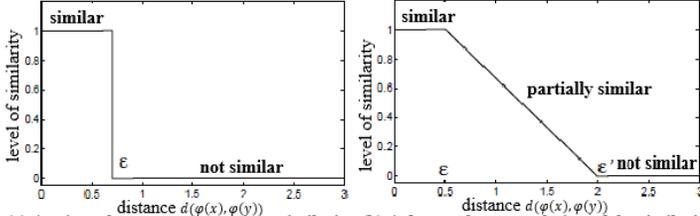

Fig.4: The transition gradual in a fuzzy tolerance relation of the similarity.

### C. Finding tolerance classes

For each visual element $x_0$ in the union of all sub-images ($x_0 \in X \cup Y$), find the tolerance classes pertaining to the tolerance relation $\cong_{\varphi,\varepsilon}$. In fact, tolerance classes are made up of the query points of consecutive neighborhoods, and then all the tolerance classes containing $x \in X$ are subsets of the neighborhood of x [25]. Finding tolerance classes is based on the Maximal Clique Enumeration (MCE) approach. This concept signifies the use of a tree structure to discover all the maximal cliques through a depth-first search, in which each call to Clique Enumerate generates a new child node. The overall idea is to find maximal cliques through a Depth-First Search where the branches are formed on the basis of candidate cliques and the back tracking takes place once a maximal clique has been discovered. This algorithm determines all the tolerance classes. The main idea behind using tolerance classes is the assumption that when we look at two images, we tend to group image elements together based on similarity to the element of interest at the point of gaze point.

### D. Computing the Tolerance Fuzzy Nearness Measure

In a tolerance space view to image correspondence, the nearness between sets of describable objects X, Y is defined by the comparison of the tolerance classes of nearly similar objects in a tolerance space that covers both images. It is meant by the nearness measure of Henry and Peters is to seek the similarity level between two variables by eliminating the existing correlation between the set of variables, called the tolerance classes. The similarity measures may be assembled into comparable classes of measures. The Tolerance Fuzzy Nearness Measure between two fuzzy sets X; Y builds on the notion that correspondent classes formed from objects in the union $Z = X \cup Y$ should be evenly divided between X and Y if these sets are similar. A Tolerance Fuzzy Nearness Measure (TFNM) is proposed here as a numerical valued crisp nearness measure obtained using a fuzzy tolerance relation. TFNM between pairs of images X, Y is defined by the following equation.

**Definition:**

Let $(O,F)$ be a perceptual system, with $\varepsilon \in R_0^+$ and $B \subset F$. Besides, let X and Y be two disjoint sets. A tolerance nearness measure between two sets X and Y is determined by:

$$TFNM_{\cong_{B,\varepsilon}}(X,Y) = 1 - \left(\sum_{\mu_C \in H_{B,\varepsilon}(Z)} |\mu_C|\right)^{-1} \cdot \sum_{\mu_C \in H_{B,\varepsilon}(Z)} |\mu_C| \frac{min(|min(\mu_C,\mu_X)|,|min(\mu_C,\mu_Y)|)}{max(|min(\mu_C,\mu_X)|,|min(\mu_C,\mu_Y)|)}$$
(14)

where $H_{\cong_{\varphi,\varepsilon}}(Z)$ is the set of fuzzy tolerance classes. Note that X and Y are pairs of images and X, Y represent sets of describable objects (visual elements) corresponding to images X,Y. When the cardinality of a fuzzy set is defined as the sum of the membership values of all the elements in a set (as defined in [36]).

**Definition:**

The Tolerance Interval-Type-2 Beta Fuzzy Nearness Measure is the average of Tolerance Interval-Type-2 Fuzzy-upper Nearness Measure and Tolerance Type-2 Fuzzy-lower Nearness Measure.

$$IT2BFNM = (IT2BFNM_{Upper} + IT2BFNM_{Lower})/2 \quad (15)$$

The best measure of similarity offers the largest number of relevant images. The measure of similarity between images is assimilated to a calculation of distance between the descriptor vector of the query image and that of an image of the base. Both the distance is small as the two images are similar.

### E. Returning query results

The system returns the result of the search in a list of ordered images according to the similarity between their descriptors and the descriptor of the query image. The effectiveness of the search is evaluated according to the number of images relevant and irrelevant to the query, found in a database: a search making it possible to find, in an image database, all the images relevant to the request, and no irrelevant image, is perfectly effective.

## V. RESULTS AND ANALYSIS

CBIR is an important application of image similarity measures. In a CBIR system the search is based on the image content (i.e. information about the feature values in images) to find similar images in a dataset. In a query by example CBIR, the query is an example image and the objective is to search in a set of images and find the ones that are similar to the given query. In this paper, we are dealing with a query by example CBIR problem. A measure is used to calculate the similarity between the query image and each image in a data set. The images are then sorted based on their similarity to the query image.

*A. DataBase*

The SIMPLIcity 1000 images dataset (available for download from [37]) is used here in a broad-domain, broad target CBIR experiment. This is a controlled test dataset and images are numbered between 0 to 999 and divided into 10 conceptually different categories (named here as target sets C0 to C9). Figure 6 displays the first 40 images in each category. Images are $384 \times 256$ pixels (dimensions). Any image from the dataset can be selected as a query image and compared to all images in the dataset.

The experiment consists of calculating the similarity measures between each query image and all 1000 images, resulting in 1; 000; 000 trials of image comparison. Subsequently, the images will be sorted based on their similarity to the query image. The experiment is performed using each one of the proposed similarity measures in previous chapters.

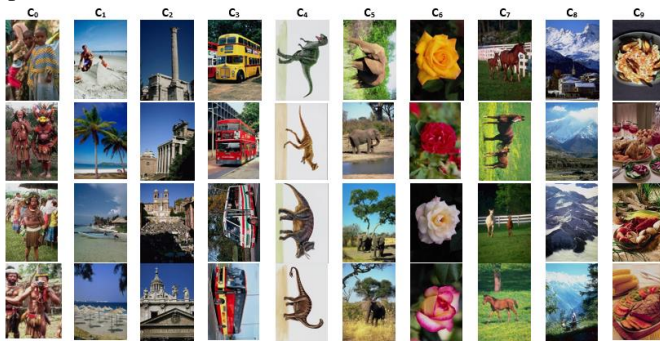

Fig. 5. Example of images of SIMPLIcity database.

In the case of the algorithms, only the results for $\varepsilon = 0.3$, since this value produces the best results that are achievable with reasonable runtime, as it was mentioned in [38]. The probe functions of the features are selected by matching the attributes chosen to describe the images of the bases and the visual requests of the users to obtain a satisfactory match. Next, the fuzzification method proceeds by three steps, the first of which is to find the membership of each object. While the second step pertain to the matching of the value of the target images with that of the source image, the third one relates to the introduction of the threshold. After that, the measurement of the degree of fuzzy nearness of all sub-images of one image is performed with the corresponding sub-image of another image using fuzzy nearness measure. Furthermore, the best similarity measure offers the greatest number of relevant images. The similarity measure between images is assimilated to a fuzzy nearness degree calculation (TFNM) between the query image and the image of the database. Both the distance is small as the two images are similar. Finally, precision and recall have been calculated for each image in the database (chosen as a query) and the values have been averaged among all queries.

**Three different methods** were used in this experiment to assess the accuracy of the image retrieval. In the first evaluation method, Precision versus recall in this example have been plotted in Fig 6-11(b). In a second method of retrieval evaluation, both precision and recall were calculated at each number of the 40 most similar images and the values of precision were plotted against recall in Fig 6-11(c). In a third evaluation method Comparison of average precision for each category between the proposed method TFNM with the results published in [20], in [39], and in [35] in table 1.

The present experimental results were achieved by the use of three approaches:

- **Near system**: is a various applications of the near set theory, and thus for measuring the perceptual nearness of objects [40].
- **Henry system**: is method of measuring perceptual nearness as the Near system, but it is based on the MCE method that seeks all classes of tolerance [38].
- **BFNSs**: (Beta Fuzzy Near Sets) our method which is based on the Near set approach hybridize with the beta function of fuzzy set approach [25].
- **IT2BFNSs**: (Interval-Type-2 Beta Fuzzy Near Sets) our method which is based on the Near set approach hybridizes with the Interval-Type-2 function of fuzzy set approach.

Some examples of experimental results obtained from the methods studied are presented.

*B. Performance Measurement for Similar Image Recovery*

To qualitatively evaluate the accuracy of the system over the 1000 image SIMPLIcity database, the best of categories of fuzzy nearness measure are selected. For each query example, the precision of the query results depending on the relevance of the image similarity is examined. In the CBIR system, it is common to use both the Recall and Precision functions to measure the performance of the system in retrieving images relevant to the query. Precision: is the percentage of the relevant images found compared to the number of all images found by the query. The precision is the number of relevant images retrieved in relation to the total number of images proposed by the search engine for a given query. The principle is that when a user queries a database, he wants the images offered in response to his query to match his expectations. Any unnecessary or irrelevant returned images are noise. The precision is opposed to this noise. If it is high, it means that few unnecessary images are offered by the system and that the latter can be considered "precise". Precision is calculated using the following formula:

$$Precision = \frac{|\{relevant\ images\} \cap \{retrieved\ images\}|}{\{retrieved\ images\}} \quad (16)$$

Recall: is the percentage of the relevant images found compared to the total number of relevant documents A perfect image search system will provide responses with accuracy and recall equal to 1 (the algorithm finds all relevant images (reminder) and makes no error).

$$Recall = \frac{|\{relevant\ images\} \cap \{retrieved\ images\}|}{\{relevant\ images\}} \quad (17)$$

In reality, the search algorithms are relatively precise, and approximately relevant. It will be possible to obtain a very precise system (for example an accuracy score of 0:99), but with poor performance (for example with a reminder of 0:2, which means that only 20% of the possible answers have been found). Similarly, an algorithm with strong recall (eg 0:99 is almost all relevant documents), but low precision (eg 0:2) will provide many erroneous images in addition to the relevant ones, and therefore will be difficult to exploit. For example, an image search system that returns all of its base images will have a 1 (but poor accuracy) reminder. While a search system that returns only the user's query will have an accuracy of 1 for a very low reminder.

Precision and Recall are interesting for a final assessment of one category. However, for larger evaluation purposes, we consider the Precision/Recall curve. This curve is the set of all the couples (Precision, Recall) for each number of images returned by the system. The curve always starts from the top left (1; 0) and ends in the bottom right (0; 1). Between this two points, the curve decreases regularly. A good Precision=Recall curve is a curve which decreases slowly, which means that at the same time, the system returns a lot of relevant images and few of them are lost.

*C. Results and Analysis*

These results present a comparison of the different approaches:
The Near system vs. Henry system vs. Beta-Fuzzy-Near system vs. type2-Beta-Fuzzy-Near system.

In the case of the algorithms, only the results for $\varepsilon = 0.3$ since this value produces the best results that are achievable with reasonable runtime, as it was mentioned in [25]. The graphs (Figure 6 to Figure 7) clearly show the difference in performance between the four systems. We presented only category 0 and 1, others category has same results. It should be born in mind that when searching images, the criterion of accuracy is paramount just like the recall elsewhere. Generally, when a user submits his request, he automatically expects precise and numerous answers at once. It thus becomes inevitable that an information search system has a compromise between the quantity and quality of images found. This imposes a compromise between accuracy and recall. Ideally, all images in the same category would be recovered before images from other categories. In this case, the accuracy would be 100% until the recall reaches 100%, at which the accuracy would fall to the number of images in the query category / number of images in the database. Therefore, our final accuracy value will be 11%. since we used 9 categories, of 100 images each. It is to be noted that only 9 categories were used since the category of images shown in Figure 5 is easy to recover and their inclusion in the test would only increase the execution time of the experiment. We observe that the base Precision / Recall curves are decreasing overall, the accuracy decreases as irrelevant images are found. This curve is the set of all the pairs (Precision, Recall) for each number of images returned by the system. The curve always starts at the top left (1.0) and ends at the bottom right (0.1). Between these two points, the curve decreases steadily. A good Precision / Recall curve is a slowly decreasing curve, which means that the system returns a lot of relevant images at the same time and few of them are lost.

The experimental results confirm that the performance of the proposed IT2BFNS technique outperforms the existing state of the art (Henry system or Near system) in CBIR. These figures reveal that IT2BFNS method leaves less mistakes than MCE-Near approach and Near system for this base. It is noteworthy that these examples represent a good illustration of the operation of our system on these bases. This can be interpreted by the specificity of the image/feature factor. We find that the result is satisfactory. Some of the curves have an acute inflection point

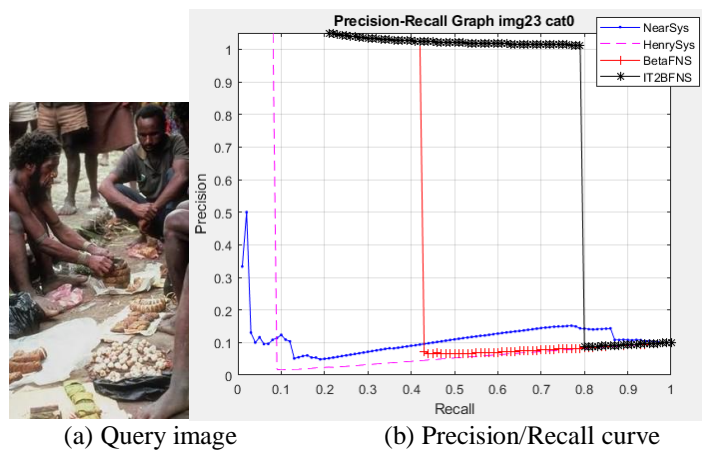

(a) Query image    (b) Precision/Recall curve

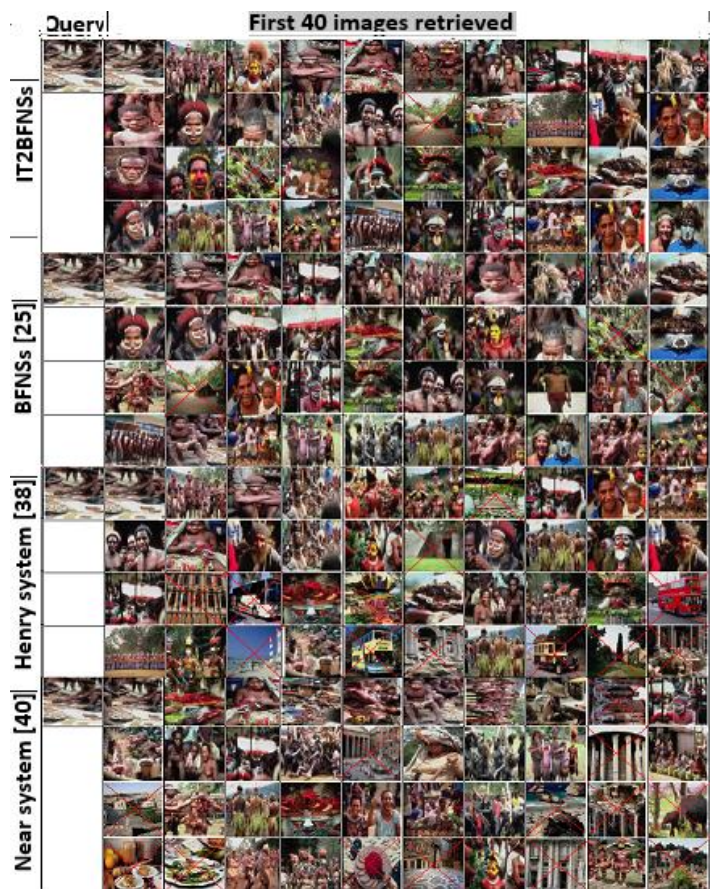

(c) First 40 images retrieved
Fig. 6. Experiment results of the best image: image 23 category 0

(see, for example in Figure 6, e = 0.8 to 9% recall in black curve, e =0.45 to 8% recall in red curve, e=0.1 to 2% recall in magenta curve and e =0.1 to 11% recall in blue curve). These points represent the location where the remaining TFNM values for a specific request become null.

The first 40 images extracted from the best search of the query are sorted and displayed according to the nearness measure in a category for each database in figures and these results affirm the improvement of the retrieval performance. We observe that these examples are a good illustration of the operation of the proposed system on these bases. This can be interpreted by the specificity of the image/feature factor.

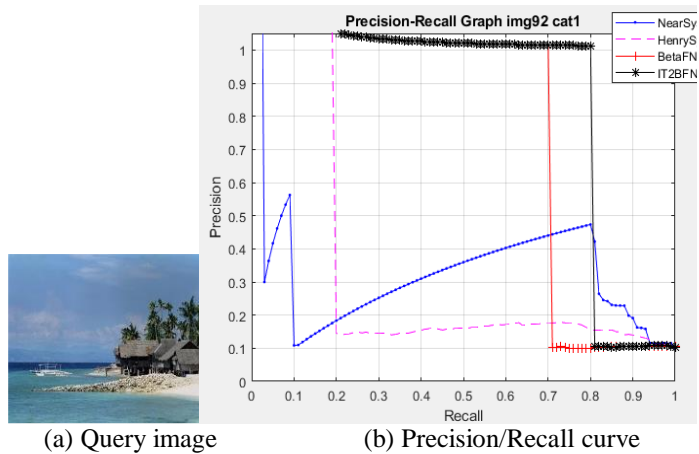

(a) Query image      (b) Precision/Recall curve

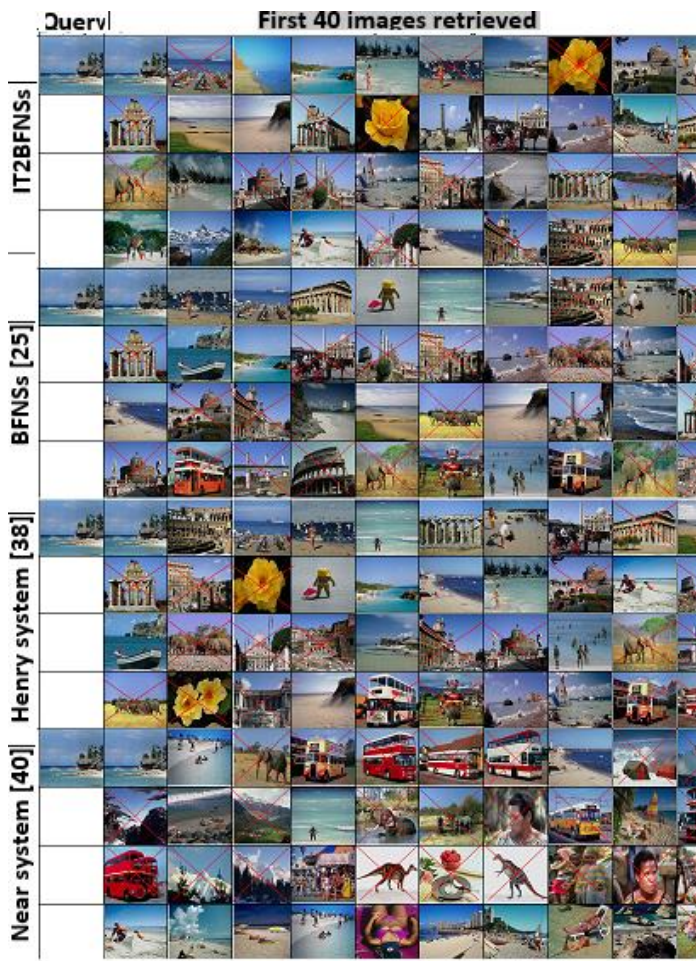

(c) First 40 images retrieved

Fig. 7. Experiment results of the best image: image 92 category 1

We find that the result is satisfactory. Eight or nine out of ten are found to be relevant images for the method IT2BFN.

Table 1 shows the average precision for all categories of images up to 100 images. The highest retrieval efficiency 100% is observed in Dinosaurs, Buses, Elephants, Horses, and Flowers. The lowest retrieval efficiency 53% is observed in mountains. The overall efficiency of the proposed approach is 100%. The performance of proposed system is compared with Nearness method in [35], and wavelet decomposition with morphological operator [20]. This results demonstrate that this method is the best compared to the other methods.

| set | Sample | Results reported in [39] | Results reported in [35] | Results reported in [20] | Results of our method TFNM |
|---|---|---|---|---|---|
| C0 |  | 76.3% | 83.8% | 80% | 88.2% |
| C1 |  | 72.5% | 73.6% | 94% | 94.06% |
| C2 |  | 86.2% | 85.9% | 81% | 89% |
| C3 |  | 92.3% | 71.4% | 99% | 100% |
| C4 |  | 100% | 100% | 100% | 100% |
| C5 |  | 74.8% | 69.05% | 100% | 100% |
| C6 |  | 89.2% | 100% | 100% | 100% |
| C7 |  | 100% | 100% | 98% | 100% |
| C8 |  | 66.8% | 58.85% | 80% | 90% |
| C9 |  | 78.7% | 80.35% | 80% | 80% |

TABLE I: COMPARISON OF AVERAGE PRECISION FOR EACH CATEGORY BETWEEN THE PROPOSED METHOD IT2BFNM WITH RESULTS PUBLISHED IN [41], IN [39], AND IN [35]

## VI. CONCLUSION

Among the many approaches used to tackle CBIR problem, the use of IT2BFNS approach is a fairly standard approach that provides satisfactory results, exhibiting some robustness to rotation, zoom, resolution change and partial occlusion. In this paper, we present an approach for image retrieval based on near set theory hybridization with Type-2 Beta Fuzzy Sets, to increase the accuracy of the correspondence. We have shown that image retrieval on type 2 beta fuzzy with near sets is more relevant than that recovering with only near sets or with beta fuzzy sets. Through this work, we show that there is a close relationship between the Near Sets Theory and Fuzzy Set Theory with various input representation models. The performance of using the Near Set approach has been proved throughout the SIMPLIcity database.

In future work, more reflection about other type of data (sound, video, ...) may be investigated.

ACKNOWLEDGEMENT

The research leading to these results has received funding from the Ministry of Higher Education and Scientific Research of Tunisia under the grant agreement number LR11ES48.